\documentclass{article}
\usepackage{spconf,amsmath,graphicx}
\usepackage{amssymb}
\usepackage{booktabs}
\usepackage{bbding}
\usepackage{float}
\usepackage{subfig}

\title{Cross-Inferential Networks \\for Source-free Unsupervised Domain Adaptation}
\name{Yushun Tang\textsuperscript{1,2*}, Qinghai Guo\textsuperscript{2\dag}, and Zhihai He\textsuperscript{1,3\dag}
}

\address{\normalsize \textsuperscript{1}Department of Electronic and Electrical Engineering, Southern University of Science and Technology, Shenzhen, China, \\ \normalsize \textsuperscript{2}Advanced Computing and Storage Laboratory, Huawei Technologies Co., Ltd., Shenzhen, China,  \normalsize \textsuperscript{3}Pengcheng Laboratory, Shenzhen, China 
\\{\tt\small tangys2022@mail.sustech.edu.cn, guoqinghai@huawei.com, hezh@sustech.edu.cn}
} 

\begin{document}
\ninept
\maketitle
\renewcommand{\thefootnote}{\fnsymbol{footnote}}
\footnotetext[1]{~Most of this work was done during Yushun Tang’s internship at Huawei.}

\footnotetext[2]{~Corresponding authors.} 
\renewcommand{\thefootnote}{\arabic{footnote}}
\begin{abstract}
One central challenge in source-free unsupervised domain adaptation (UDA) is the lack of an effective approach to evaluate the prediction results of the adapted network model in the target domain. To address this challenge, we propose to explore a new method called \textit{cross-inferential networks} (CIN). Our main idea is that, when we adapt the network model to predict the sample labels from encoded features, we use these prediction results to construct new training samples with derived labels to learn a new examiner network that performs a different but compatible task in the target domain.
Specifically, in this work, the base network model is performing image classification while the examiner network is tasked to perform relative ordering of triplets of samples whose training labels are carefully constructed from the prediction results of the base network model. Two similarity measures, cross-network correlation matrix similarity and attention consistency, are then developed to provide important guidance for the UDA process. Our experimental results on benchmark datasets demonstrate that our proposed CIN approach can significantly improve the performance of source-free UDA.
\end{abstract}

\section{Introduction}
\label{sec-intro}

The performance of deep neural networks  tends to degrade significantly when the training samples and test samples are from different environments, which is often referred to as the domain drift problem~\cite{pan2009survey}. 
To address the performance degradation problem,
various transfer learning methods have recently been developed~\cite{pan2009survey,wang2018deep,zhuang2020comprehensive,zuo2023dual}.
Unsupervised Domain Adaptation (UDA)  aims to transfer knowledge learned from the labeled datasets in  the source domain to new unlabeled datasets in the target domain.
Recently developed methods for UDA include statistical distribution matching and generative adversarial network (GAN) methods to minimize domain discrepancy~\cite{sun2016return,long2017deep,peng2019moment,ganin2015unsupervised,pei2018multi}. They have achieved very promising performance thanks to the access to source data. In real-world applications, it is often not possible to access the source-domain samples due to data privacy issues or transmission difficulties.
Under this circumstance, source-free UDA, which does not need to access the source-domain samples, is highly desirable  in practice and emerges as a new and very important research topic~\cite{liang2020we}.
Recently, several source-free UDA methods~\cite{liang2020we,Yang_2021_ICCV,wang2021tent,tang2023neuro} have been developed based on information maximization, pseudo labels, GAN, etc.
These methods attempt to establish rules and procedures to estimate the pseudo labels for samples and the corresponding loss functions in the target domain using the network model received from the source domain. 

\begin{figure}[!t]
\centering
\setlength{\abovecaptionskip}{0.1cm}
\setlength{\belowcaptionskip}{-0.4cm}
\includegraphics[width = 0.35\textwidth]{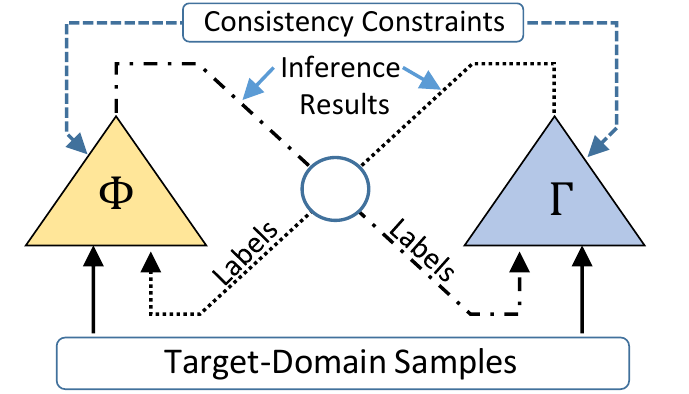}
\caption{Illustration of the proposed idea of cross-inferential networks for source-free UDA.}
\label{fig:idea}
\end{figure}

One central challenge in source-free UDA is the lack of an effective approach to evaluate the prediction results of the adapted network model.
Our main idea is that, to evaluate the  accuracy of the inference output of the base network $\Phi$, we use this inference output to construct new training samples and labels to learn a cross examiner network $\Gamma$ and watch its response and correlation with the base network $\Phi$. We hypothesize that, if the predicted results of  network $\Phi $ are accurate, the learned cross examiner network $\Gamma$ should produce fairly accurate estimations and its inference process should also exhibit strong correlation with the original inference process of the base network. 
We hope that this correlation analysis will provide important feedback information about the prediction performance of the base network and important guidance for the source-free UDA. 

Certainly, the cross examiner $\Gamma$ should be performing a different task, otherwise it will not be able to provide extra feedback or guidance information. In the meantime, its learning task should be compatible with the original task of the base network. Otherwise, it will not be able to provide helpful feedback or guidance. 
Specifically, in this work, the base network model is performing classification while the examiner network is tasked to perform relative ordering of triplets of samples whose training labels are  constructed from the prediction results of the network model.
Two similarity measures, cross-network correlation matrix similarity and attention consistency, are then developed to provide important guidance for the UDA process. 

Our \textbf{major contributions} can be summarized as follows: (1) We construct a cross examination module to provide external guidance for the source-free UDA process. These two networks, although much different in their learning tasks are totally compatible with each other in their learning processes. (2) We develop two types of cross-network consistency, correlation matrix consistency and attention consistency, to provide mutual guidance between the base network and the examiner network during their adaptation processes. (3) Our experimental results on benchmark datasets demonstrate that our proposed cross-inferential networks approach can significantly improve the source-free UDA performance.

\section{Methodology}
In this section, we present our new method of cross-inferential networks (CIN) for source-free UDA.

\subsection{Constructing the Examiner Network}
\label{sec-rrn}

We consider the task of source-free UDA where only pre-trained source models and unlabelled target samples are used. In particular, we consider a $K$-way classification task, where the source and target domains share the same label space. The source-domain training dataset has $N_s$ labeled samples $\{(x_s^i, y_s^i)|1\le i\le N_s\}$ where $x_s^i$ belongs to the source sample set $\mathcal{X}_s$ and $y_s^i$ belongs to the label set $\mathcal{Y}$. 
The target domain has $N_t$ unlabeled data $\{x_t^j\in \mathcal{X}_t|1\le j\le N_t\}$ with the same label set $\mathcal{Y}$. The goal of the source-free UDA is to predict the label of $\{x_t^j\}$ using the adapted model transmitted from the source domain.

\vspace{-0.2cm}
\begin{figure}[!ht]
\centering
\setlength{\abovecaptionskip}{0cm}
\setlength{\belowcaptionskip}{-0.4cm}
\includegraphics[width = 0.4\textwidth]{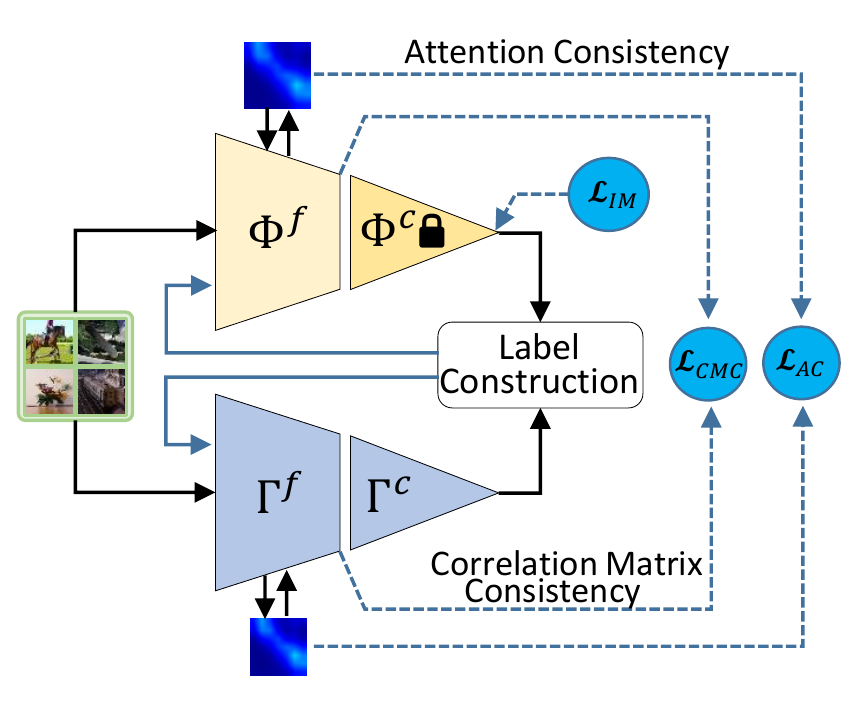}
\caption{An overview of the proposed CIN method. The base network is trained on the source domain. Then the base network is transferred to the target domain. Finally, The base network and the cross examiner network are adapted together in a collaborative manner. The lock symbol means the classifier head is fixed in the adaptation process.}
\label{fig:framework}
\end{figure}

As illustrated in Figure \ref{fig:framework}, the base network $\mathbf{\Phi}_s=\mathbf{\Phi}_s^f\oplus\mathbf{\Phi}_s^c$ is an image classification model trained on the source-domain that needs to be adapted to the target domain. 
Here, $\mathbf{\Phi}_s^f$ is the feature extractor and 
$\mathbf{\Phi}_s^c$ is the classification head. The subscript $s$ represents the source domain.
In this work, we use SHOT \cite{liang2020we} as the baseline method to perform this adaptation. 
Our proposed CIN method is implemented on top of the SHOT method, aiming to improve its performance.
In the target domain, the classification head $\mathbf{\Phi}_t^c$ remains the same as the source domain, i.e., $\mathbf{\Phi}_t^c = \mathbf{\Phi}_s^c$.
The UDA algorithm only adapts the feature encoder 
$\mathbf{\Phi}_t^f$.

To provide external guidance for this source-free UDA process, we construct an  examiner network  $\mathbf{\Gamma}_t=\mathbf{\Gamma}_t^f\oplus \mathbf{\Gamma}_t^c$ in the target domain. It performs relative ordering between images:
given an anchor image $F_a$ and two other images $F_b$ and $F_c$, $\mathbf{\Gamma }_t$ needs to determine which image, either $F_b$ or $F_c$, is closer to the anchor image $F_a$.
Let us introduce the following notation: if image $F_b$ is closer to $F_a$ than $F_c$, we have:
\begin{equation}
    \mathbf{\Gamma }_t(F_b\prec F_c | F_a) = 0.
\end{equation}
Otherwise, 
\begin{equation}
    \mathbf{\Gamma }_t(F_b\prec F_c | F_a) = 1.
\end{equation}
The examiner network is trained on the target domain with high-confidence pseudo labels.
Specifically, we first train the base network which consists of the feature network and the classifier head. 
Once trained, $\mathbf{\Phi}_s$ is adapted to the target domain in a source-free manner and becomes $\mathbf{\Phi}_t$. 
We then train the examiner network module $\mathbf{\Gamma}_t$ with the high-confidence pseudo labels based on the adapted classification network. 
In Section \ref{sec-UDA} and \ref{sec: collab}, we will explain how these two networks are further jointly adapted to the target domain in a collaborative manner based on cross-network consistency.

\subsection{Cross-Inferential Adaptation in the Target Domain }
\label{sec-UDA}
In this section, we explain how the base network $\mathbf{\Phi}_s$ is adapted to the target domain and how the examiner network $\mathbf{\Gamma}_t$ is trained.

 \textbf{(1) Adaptation of the base network.}  We use the SHOT method~\cite{liang2020we} as our baseline method to adapt the base network. Specifically, it fixes the classification module. The classification module  $\mathbf{\Phi}^c_t$ in the target domain remains the same as the original classification module in the source domain $\mathbf{\Phi}^c_s$. It only updates the feature extraction module $\mathbf{\Phi}^f_t$ in the target domain. During this learning process, the following information maximization (IM) loss~\cite{krause2010discriminative,hu2017learning} $\mathcal{L}_{\mathrm{IM}}$ is used to adapt the  target feature extractor $\mathbf{\Phi}^f_t$:
\begin{equation}
    \mathcal{L}_{IM} = 
     -\mathbb{E}_{\{x_t\}} \sum_{k=1}^{K} \delta_k[\mathbf{\Phi}_t(x_t)]  \mathrm{log}_2 \delta_k [\mathbf{\Phi}_t(x_t)] 
     + \sum_{k=1}^{K} \hat{p}_k \mathrm{log}_2 \hat{p}_k, 
\end{equation}
where $\hat{p}=\mathbb{E}_{\{x_t\}} [\delta (\mathbf{\Phi}_t(x_t))]$ is the mean output embedding of the mini-batch and $\delta$ represents softmax.

\textbf{(2) Training of the examiner network.} 
The input to the examiner network includes three images, the anchor image $F_a$ and two images $F_b$ and $F_c$ for relative ordering. 
Let $y(F_b)$ be the label for image $F_b$. We construct the training set for the examiner network based on the following rules:
\begin{equation}\label{eq_RRN_rule}
\small
{
    \!\!\left\{
    \!\!\!\begin{array}{ll}
    \!\mathbf{\Gamma}_t(F_b\prec F_c | F_a)=0, & if \ y(F_b) = y(F_a),\ y(F_c)\neq y(F_a);\\
    \!\mathbf{\Gamma}_t(F_b\prec F_c | F_a)=1, & if \ y(F_b) = y(F_a),\ y(F_c) = y(F_a). 
    \end{array}
      \right.
}
\end{equation}
To train the examiner network model $\mathbf{\Gamma}_t$ in the target domain, the following binary cross-entropy loss is used:
\begin{equation}\label{eq_bnloss}
    \!\!\!\!\mathcal{L}_{EN} = - \mathbb{E}_{F_a, F_b, F_c \in \mathcal{X}_t} \sum_{k=1}^{2} q_k\cdot \mathrm{log}_2 \delta_k[\mathbf{\Gamma}_t(F_b\prec F_c | F_a)],
\end{equation}
where $\delta_k[\cdot]$ is $k$-th element of the vector
and  $q_k$ is $k$-th element of the binary  label vector. 
Since there is no supervised label in the target domain, we first use the adapted base network $\mathbf{\Phi}_t$ to construct high-confidence training samples in the target domain for the examiner network. 
Specifically, let $Z^n=[z_1^n, z_2^n, \cdots, z_K^n]$ be the softmax output of network $\mathbf{\Gamma}_t$ for input image $I_n$ from the target domain. We define its confidence level based on its entropy:
\begin{equation}
    \mathcal{H}(Z^n) = - \sum_{k=1}^K z_k^n\cdot \log z_k^n.
    \label{eq_entropy}
\end{equation}
An image sample is said to be high-confidence if its softmax output is close to one-hot vector, or its entropy is smaller than a given threshold. 
Using these high-confidence samples from different image classes, we can construct the training samples for the examiner network according to Eq. \eqref{eq_RRN_rule}. 
Empirically, we observe that the \textit{curriculum sampling} method~\cite{li2021imbalanced} provides us relatively accurate estimation of the pseudo image labels. 
Initially, we sort all the target samples by Eq. \ref{eq_entropy} and use half of the samples with high confidence to construct the training set for the examiner network $\mathbf{\Gamma}_t$. The number of the training set is gradually increased to include all target samples as the accuracy of the base network improves.

\subsection{Cross-Network Constraints for Collaborative Adaptation} 
\label{sec: collab}

As discussed in Section \ref{sec-intro}, the examiner network is used to provide external guidance for the base network. In this work, this guidance is achieved by the following two cross-network consistency losses: (A) correlation matrix consistency, and (B) attention consistency. 

\begin{figure}[htbp]
\centering
\setlength{\abovecaptionskip}{0.1cm}
\includegraphics[width = 0.45\textwidth]{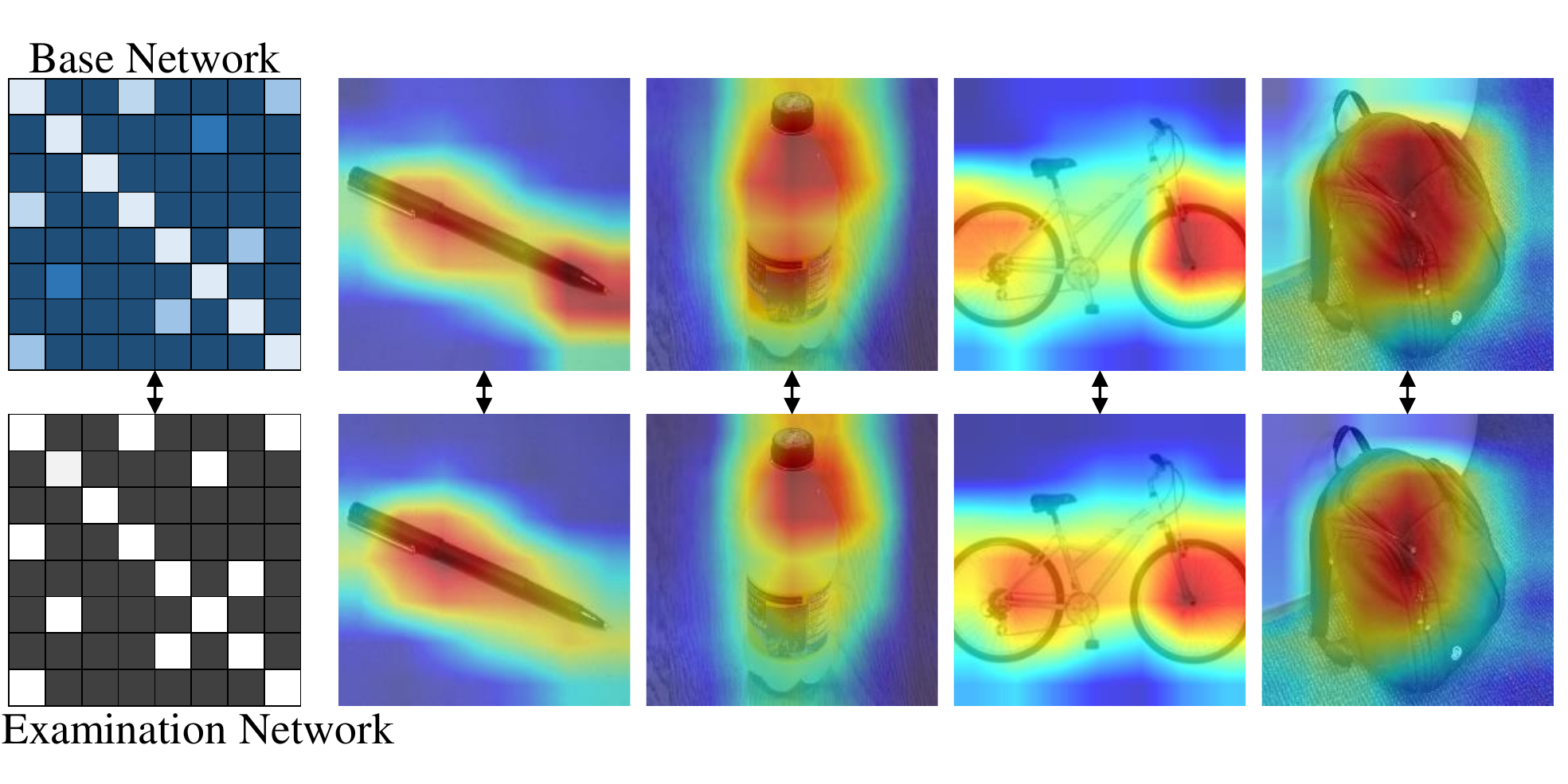}
\caption{The correlation matrix consistency (left) and attention consistency (right). In the correlation matrix, higher brightness in a cell means higher similarity between two corresponding samples.}
\label{fig:consistency}
\end{figure}

\textbf{(A) Correlation Matrix Consistency.} As defined in the above section, the examiner network is tasked to analyze the similarity within each triplet of images. It provides a measurement of the correlation between two image samples through a reference sample. Specifically, given two images $F_a$ and $F_b$, let $\Omega(F_a)$ be the augmentation of image $F_a$, we compute the correlation between $F_a$ and $F_b$ through the reference of $\Omega(F_a)$ by the following output of the examiner network: 
\begin{equation}
    \mathcal{C}_{\Gamma_t}(F_a, F_b) = \mathbf{\Gamma}_t(F_b \prec \Omega(F_a) | F_a).
\end{equation}
For all pairs of images from the mini-batch, we obtain  their correlation matrix, denoted by 
$[\mathcal{C}_{\Gamma_t}(F_a, F_b)]_{M\times M}$, where $M$ is the number of images in the mini-batch.
In the meantime, using the features from the base network, we can also compute the correlation matrix $\mathcal{C}_{\Phi_t}(F_a, F_b)$ for all images in the mini-batch. 
Specifically, we use the cosine similarity between their features $\mathbf{f}_a$ and $\mathbf{f}_b$ to define their correlations:
\begin{equation}
    \mathcal{C}_{\Phi_t}(F_a, F_b) = \frac{\mathbf{f}_a\cdot \mathbf{f}_b}{||\mathbf{f}_a||\times ||\mathbf{f}_b||}.
\end{equation}
To provide external guidance, we require that the correlation matrix $\mathcal{C}_{\Phi_t}(F_a, F_b)$ obtained by the base network should be close to the correlation matrix $\mathcal{C}_{\Gamma_t}(F_a, F_b)$ obtained by the examiner network. Figure \ref{fig:consistency} (top) shows the correlation matrices for eight images obtained by the base network and the examiner network modules.
This leads to the following correlation matrix consistency loss:
\begin{equation}
    \mathcal{L}_{CMC} = ||\mathcal{C}_{\Gamma_t}(F_a, F_b) - \mathcal{C}_{\Phi_t}(F_a, F_b) ||_2.
\end{equation}

\textbf{(B) Attention Consistency.}
Another cross-network consistency is the attention consistency. This is motivated by the following observation: when the base network and the examiner network are examining the same image, the attention maps produced by them should be similar, i.e., the spatial distributions of salient features should remain the same across these two networks. 
For example, Figure \ref{fig:consistency} (bottom) shows four examples of attention weights obtained from the base network (top) and the examiner network (bottom).  
Motivated by this, we incorporate an attention module into these two networks. 
Let $\mathcal{A}_{\Phi_t}(F)$ and $\mathcal{A}_{\Gamma_t}(F)$ be the attention weights of image $F$ from these two networks. 
The attention consistency is defined as:
\begin{equation}
    \mathcal{L}_{AC} = ||\mathcal{A}_{\Gamma_t}(F) - \mathcal{A}_{\Phi_t}(F)||_2.
\end{equation}

On top of the information maximization loss 
$\mathcal{L}_{IM}$, the combination of the correlation matrix consistency and the attention consistency is used to provide mutual guidance when training these two networks, which leads to the following loss function:
\begin{equation}
    \mathcal{L} = \mathcal{L}_{IM} + \lambda_1 \cdot \mathcal{L}_{CMC} + \lambda_2 \cdot \mathcal{L}_{AC},
\label{eq-loss}
\end{equation}
where $\lambda_1$, $\lambda_2$ are weighting parameters which are setting to $10$.

\textbf{Pre-training the examiner network in the source domain.} 
In this work, the examiner network is tasked to perform relative ordering between images and is added to the base network. 
If we choose to pre-train this examiner network module in the source domain and then send this network module to the target domain along with the baseline base network for joint UDA, the performance can be further improved. The corresponding algorithm is referred to as 
\textit{CIN (Pre-Trained)}. Otherwise, the original algorithm is referred to as \textit{CIN}.

\begin{table}[ht]
\setlength{\abovecaptionskip}{0cm}
\caption{ Classification accuracy(\%) on \textbf{Office-31} for Source-free UDA. The best and second-best results are shown in bold and underscore respectively in the table.}
\label{table:office-31}
\begin{center}
\resizebox{0.48\textwidth}{!}{
\begin{tabular}{lcccccc|r}
\toprule
Methods  & A$\rightarrow$D & A$\rightarrow$W & D$\rightarrow$A & D$\rightarrow$W & W$\rightarrow$A & W$\rightarrow$D & Avg.\\
\midrule
Source-only~\cite{liang2020we}    & 80.8 & 76.9 & 60.3 & 95.3 & 63.6 & 98.7 & 79.3\\ 
SHOT~\cite{liang2020we}    & 94.0 & 90.1 & 74.7 & 98.4 & 74.3 & \underline{99.9} & 88.6\\ 
3C-GAN~\cite{li2020model}   & 92.7 & \underline{93.7} & 75.3 & 98.5 & $\boldsymbol{77.8}$ & 99.8 & 89.6\\ 
CPGA~\cite{qiu2021source}   & 94.4 & $\boldsymbol{94.1}$ & \underline{76.0} & 98.4 & 76.6 & 99.8 & 89.9\\  
HCL~\cite{huang2021model} & 94.7 & 92.5 &75.9 & 98.2 &\underline{77.7} &$\boldsymbol{100.0}$ &89.8\\
DIPE~\cite{Wang_2022_CVPR}   & $\boldsymbol{96.6}$ & 93.1 & 75.5 & 98.4 & 77.2 & 99.6 & $\boldsymbol{90.1}$\\
\noalign{\smallskip}
\hline
\noalign{\smallskip}
$\textbf{CIN}$   & 95.6 & 91.8 & $\boldsymbol{76.3}$ & \underline{98.6} & 76.7  & $\boldsymbol{100.0}$ & 89.8\\
$\textbf{CIN}$ (\textit{Pre-trained})  & \underline{95.8} & 93.1 & 75.7 & $\boldsymbol{98.7}$ & 76.8 & 99.8 & \underline{90.0}\\
\bottomrule
\end{tabular}}
\end{center}
\end{table}
\setlength{\tabcolsep}{1.4pt}

\setlength{\tabcolsep}{2pt}
\begin{table*}[htbp]
\setlength{\abovecaptionskip}{0cm}
\begin{center}
\caption{ Classification accuracy(\%) on \textbf{Office-Home} for Source-free UDA. 
}
\label{table:office-home}
\resizebox{0.8\linewidth}{!}{
\begin{tabular}{lcccccccccccc|r}
\toprule
Methods  & Ar$\rightarrow$Cl & Ar$\rightarrow$Pr & Ar$\rightarrow$Rw & Cl$\rightarrow$Ar & Cl$\rightarrow$Pr & Cl$\rightarrow$Rw & Pr$\rightarrow$Ar & Pr$\rightarrow$Cl & Pr$\rightarrow$Rw & Rw$\rightarrow$Ar & Rw$\rightarrow$Cl & Rw$\rightarrow$Pr & Avg.\\
\midrule
Source-only~\cite{liang2020we}    & 44.6 & 67.3 & 74.8 & 52.7 & 62.7 & 64.8 & 53.0 & 40.6 & 73.2 & 65.3 & 45.4 & 78.0 & 60.2\\ 
SHOT~\cite{liang2020we}   & 57.1 & 78.1 & 81.5 & 68.0 & 78.2 & 78.1 & 67.4 & 54.9 & 82.2 & 73.3 & 58.8 & 84.3 & 71.8\\ 
G-SFDA~\cite{Yang_2021_ICCV}   & 57.9 & 78.6 & 81.0 & 66.7 & 77.2 & 77.2 & 65.6 & 56.0 & 82.2 & 72.0 & 57.8 & 83.4 & 71.3\\
CPGA~\cite{qiu2021source}    & \underline{59.3} & 78.1 & 79.8 & 65.4 & 75.5 & 76.4 & 65.7 & \underline{58.0} & 81.0 & 72.0 & \underline{64.4} & 83.3 & 71.6\\              
HCL~\cite{huang2021model}   & $\boldsymbol{64.0}$ & 78.6 & $\boldsymbol{82.4}$ & 64.5 & 73.1 & $\boldsymbol{80.1}$ & 64.8 & $\boldsymbol{59.8}$ & 75.3 & $\boldsymbol{78.1}$ & $\boldsymbol{69.3}$ & 81.5 & 72.6\\
DIPE~\cite{Wang_2022_CVPR}  & 56.5 & 79.2 & 80.7 & $\boldsymbol{70.1}$ & 79.8 & \underline{78.8} & \underline{67.9} & 55.1 & $\boldsymbol{83.5}$ & 74.1 & 59.3 & \underline{84.8} & 72.5\\
\noalign{\smallskip}
\hline
\noalign{\smallskip}
$\textbf{CIN}$   & 57.8 & \underline{79.3} & \underline{81.9} & \underline{68.1} & \underline{80.8} & \underline{78.8} 	& $\boldsymbol{68.1}$ 	& 56.5 	& \underline{82.4} 	& \underline{75.1} 	& 61.7 	& $\boldsymbol{85.0}$ 	& $\boldsymbol{73.0}$ \\ 
$\textbf{CIN}$ (\textit{Pre-trained})   & 57.6 & $\boldsymbol{79.5}$ & 81.8 & 67.8 & $\boldsymbol{81.0}$ & 78.7 & \underline{67.9} & 56.5 & 82.3 & 74.8 & 61.9 & \underline{84.8} & \underline{72.9}\\ 	
\bottomrule
\end{tabular}
}
\end{center}
\end{table*}
\setlength{\tabcolsep}{1.4pt}

\begin{figure*}[htbp]
    \centering
    \begin{minipage}{0.65\linewidth}
		\centering        
\setlength{\abovecaptionskip}{0cm}
\captionof{table}{Classification accuracy(\%) on \textbf{VisDA-2017} for Source-free UDA.}
\label{table:visda}
\resizebox{\textwidth}{!}{
\begin{tabular}{lcccccccccccc|r}
\toprule
Methods & plane & bcycl & bus & car & horse & knife & mcycl & person & plant & sktbrd & train & truck & Avg.\\			
\midrule
Source-only~\cite{liang2020we}    & 60.9 & 21.6 & 50.9 & 67.6 & 65.8 & 6.3 & 82.2 & 23.2 & 57.3 & 30.6 & 84.6 & 8.0 & 46.6\\ 
SHOT~\cite{liang2020we}    & 94.3  & $\boldsymbol{88.5}$  & 80.1  & 57.3  & 93.1  & 94.9  & 80.7  & 80.3  & 91.5  & 89.1  & 86.3  & 58.2  & 82.9\\ 
3C-GAN~\cite{li2020model}    & 94.8  & 73.4  & 68.8  & \underline{74.8}  & 93.1  & 95.4  & 88.6  & \underline{84.7}  & 89.1  & 84.7  & 83.5  & 48.1  & 81.6\\
G-SFDA~\cite{Yang_2021_ICCV}   & $\boldsymbol{96.1}$ & \underline{88.3} & $\boldsymbol{85.5}$ & 74.1 & $\boldsymbol{97.1}$ & 95.4 & 89.5 & 79.4 & $\boldsymbol{95.4}$ & $\boldsymbol{92.9}$ & \underline{89.1} & 42.6 & \underline{85.4}\\
CPGA~\cite{qiu2021source}  & 94.8 &  83.6 &  79.7 &  65.1 &  92.5 &  94.7 &  \underline{90.1} &  82.4 &  88.8 &  88.0 &  88.9 &  \underline{60.1} & 84.1\\
HCL~\cite{huang2021model}   & 93.3 & 85.4 & 80.7 & 68.5 & 91.0 & 88.1 & 86.0 & 78.6 & 86.6 & 88.8 & 80.0 & 74.7 & 83.5\\
DIPE~\cite{Wang_2022_CVPR}  & \underline{95.2} & 87.6 & 78.8 & 55.9 & \underline{93.9} & 95.0 & 84.1 & 81.7 & 92.1 & 88.9 & 85.4 & 58.0 & 83.1\\
\noalign{\smallskip}
\hline
\noalign{\smallskip}
$\textbf{CIN}$   & 94.9 & 83.9 & 78.5 & 65.3 & 93.4 & \underline{95.9} & 89.3 & 80.1 & 91.6 & \underline{91.1} & 86.4 & 58.3 & 84.1\\
$\textbf{CIN}$ (\textit{Pre-trained})    & 93.6 	 & 81.5  & 	\underline{83.6}  & 	$\boldsymbol{76.5}$  & 	93.8  & $\boldsymbol{96.8}$  & 	$\boldsymbol{93.5}$  & 	$\boldsymbol{85.9}$  & 	\underline{93.5}  & 	\underline{91.1}  & 	$\boldsymbol{90.1}$  & 	$\boldsymbol{60.3}$ 
& $\boldsymbol{86.7}$\\
\bottomrule
\end{tabular}
}
\end{minipage}
        \hfill
	\begin{minipage}{0.33\linewidth}
		\centering
  \setlength{\abovecaptionskip}{0.1cm}
    \subfloat[Source-only]{\includegraphics[width = 0.5\textwidth]{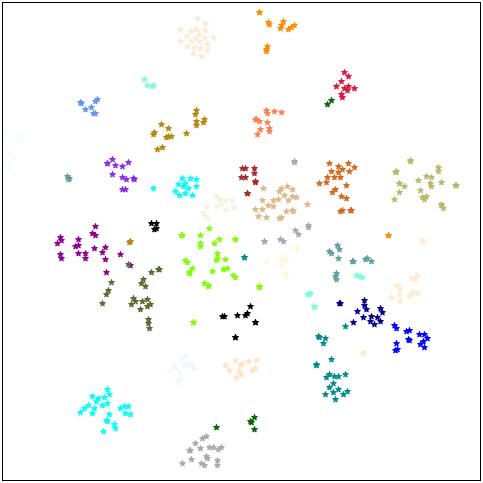}}
    \subfloat[Ours]{\includegraphics[width = 0.5\textwidth]{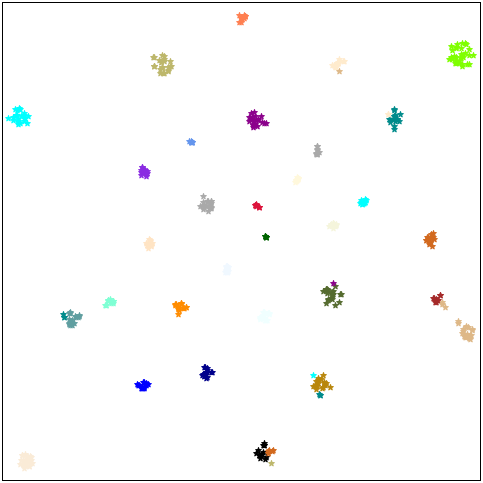}}
\caption{The t-SNE visualizations of target features in task A$\rightarrow$D of Office-31. (a) Source-only; (b) Ours. 
}
\label{fig:tsne}
	\end{minipage}
\end{figure*}

\section{Experiments}
In this section, we conduct experiments on multiple domain adaptation benchmark datasets to evaluate the performance of our method.
For each task, we only use source data to obtain the pre-trained source models. 

\vspace{-0.2cm}
\subsection{Experimental Settings}

\textbf{(A) Benchmark Datasets.}
We evaluate our method among the following three popular benchmark datasets for UDA. (1) \textbf{Office-31}~\cite{saenko2010adapting} is a standard domain adaptation benchmark, encompassing images from three distinct domains: Amazon (\textbf{A}), DSLR (\textbf{D}), and Webcam (\textbf{W}). Each domain comprises a set of $31$ classes within the office environment and contains $2817$, $498$, and $795$ samples, respectively. 
Following~\cite{saenko2010adapting}, we evaluate all six domain adaptation tasks: A$\rightarrow$D, A$\rightarrow$W, D$\rightarrow$A, D$\rightarrow$W, W$\rightarrow$A, W$\rightarrow$D. 
(2) Office-Home dataset~\cite{venkateswara2017deep} presents a challenging medium-sized benchmark comprising images from four diverse domains: artistic images (\textbf{Ar}), clip art (\textbf{Cl}), product images (\textbf{Pr}), and real-world images (\textbf{Rw}).
It contains $15,500$ images and each of these four domains has $65$ categories. 
Following~\cite{venkateswara2017deep}, we consider $12$ transfer tasks for all combinations of source and target domains. 
(3) \textbf{VisDA-2017}~\cite{peng2017visda} is a more challenging large-scale benchmark designed for synthetic-to-real domain adaptation, comprising a set of $12$ classes. The source domain comprises $152,397$ synthetic images, while the target domain consists of $55,388$ real-world images. 

\textbf{(B) Comparison Methods.}
We compare our method with recent methods in the literature from the following categories: (1) source-only; 
(2) source-free UDA methods which include 
SHOT~\cite{liang2020we},
3C-GAN~\cite{li2020model}, 
G-SFDA~\cite{Yang_2021_ICCV}, 
CPGA~\cite{qiu2021source},
HCL~\cite{huang2021model},
 and DIPE~\cite{Wang_2022_CVPR}.

\textbf{(C) Implementation details.}
Following existing work in the literature, we utilize the ResNet50 architecture, pre-trained on ImageNet, as the backbone for Office-31 and Office-Home datasets. For VisDA-2017, we employ the ResNet101 architecture with the same pre-training.
For the examiner network, we use a classification head with $256$ units hidden layer to generate the correlation matrix. For the attention, the ECANet~\cite{Wang2020ECANetEC} block is used after \textit{layer\ 4} in the ResNet.

\vspace{-0.2cm}
\subsection{Performance Results}
The classification accuracy in the target domains for source-free UDA is reported in Tables~\ref{table:office-31}-\ref{table:visda}, where the comparison methods' results are directly cited from their respective original papers.
Table~\ref{table:office-31} compares the classification accuracy of our proposed method against recent UDA methods on the Office-31 datasets.
Table~\ref{table:office-home} shows the performance comparison results on the  Office-Home dataset. 
We can see that our method achieves the best average accuracy when compared to other source-free UDA methods.
Very encouraging results are also obtained on the VisDA-2017 dataset, as shown in Table~\ref{table:visda}.  
Our method improves the average accuracy by $\textbf{3.8\%}$ over the baseline SHOT, which is quite significant.

\vspace{-0.2cm}
\subsection{Ablation Studies}

\textbf{(A) Contributions of algorithm components.}
We conduct ablation study with source-free UDA tasks on the three datasets to investigate the contributions of different components in our method. From Table~\ref{table:ablation_visDA}, we can see that all major algorithm modules, namely the attention consistency and the correlation matrix consistency contribute significantly to the overall performance.

\textbf{(B) Feature visualization.}
Fig.~\ref{fig:tsne} visualizes the embedded features on the task A$\rightarrow$D of Office-31 using t-SNE. Before adaptation, as shown in Fig.~\ref{fig:tsne}(a), the target domain features generated by the source model scatter around. 
However, our method is able to aggregate the image features into more compact and accurate clusters, as shown in Fig.~\ref{fig:tsne}(b).

\begin{table}[!ht]
\begin{center}
\setlength{\abovecaptionskip}{0cm}
\caption{Contributions of algorithm components.}
\label{table:ablation_visDA}
\resizebox{0.48\textwidth}{!}{
\begin{tabular}{lccclclclclcl}
\toprule\noalign{\smallskip}
Methods  & Office-31  & Office-Home & VisDA-2017\\
\noalign{\smallskip}
\hline
\noalign{\smallskip}
$\textbf{Our CIN Algorithm}$ (Full version)  & $\boldsymbol{90.0}$ & $\boldsymbol{72.9}$ & $\boldsymbol{86.7}$\\
\quad w/o Attention Consistency    &89.2 &72.7 & 84.6\\
\quad w/o Correlation Matrix Consistency    &89.3 &72.7 & 84.4\\
Baseline Method (SHOT)  &88.6 &71.8  & 82.9\\
\bottomrule
\end{tabular}
}
\end{center}
\end{table}
\setlength{\tabcolsep}{1.4pt}
\vspace{-0.6cm}

\section{Conclusion}
In this work, we developed a new approach for source-free UDA.
We incorporate a dual module which performs relative ordering for a triplet of images into the base network to guide its UDA process. 
These two networks, although much different in their learning tasks, are totally compatible with each other in their learning processes. 
They analyze each other's outputs and establish self-matching constraints between them  in the form of cross-network correlation and attention consistency. These constraints provide important external guidance from a new perspective for the base network during its adaptation process. 
Experimental results on benchmark datasets demonstrate that our proposed cross-inferential approach can significantly improve the source-free UDA performance.

\bibliographystyle{IEEEbib}
\bibliography{refs}

\end{document}